\documentclass{article}

\usepackage{graphicx}
\usepackage{amsmath}
\usepackage{amssymb}
\usepackage{booktabs}
\usepackage{tikz}
\usetikzlibrary{positioning,fit,shapes.geometric,arrows.meta,calc,backgrounds}
\usepackage{natbib}
\usepackage{caption}
\usepackage{algorithm}
\usepackage{algorithmic}
\usepackage{newfloat}
\usepackage{listings}
\usepackage[margin=1in]{geometry}
\usepackage{hyperref}

\DeclareCaptionStyle{ruled}{labelfont=normalfont,labelsep=colon,strut=off}
\floatstyle{ruled}
\newfloat{listing}{tb}{lst}{}
\floatname{listing}{Listing}

\title{RAGA: Reading-And-Graph-building-Agent for Autonomous Knowledge Graph Construction and Retrieval-Augmented Generation}
\author{Chengrui Han \quad Zesheng Cheng \\
Qingdao University}
\date{}

\begin{document}

\maketitle

\begin{abstract}
Existing LLM-driven knowledge graph (KG) construction methods predominantly employ stateless batch processing pipelines, exhibiting structural deficiencies in cross-chunk semantic relation capture, entity disambiguation, and construction process interpretability. These limitations undermine KG quality, retrieval precision, and deployment trust in high-stakes domains.

We propose RAGA (Reading And Graph-building Agent), an LLM-based autonomous KG construction and retrieval fusion framework. RAGA provides an atomic toolset supporting full KG lifecycle CRUD operations and embeds a ``Read--Search--Verify--Construct'' cognitive constraint into a ReAct tool loop. A KG-vector synchronization mechanism enables hybrid symbolic-vector retrieval, while evidence-anchored verification links every knowledge entry to its source text for auditable provenance.

Preliminary experiments on a subset of the QASPER scientific QA dataset indicate that RAGA's fusion retrieval outperforms zero-shot baselines, with KG integration providing measurable gains in both answer and evidence quality. The framework design and experimental baseline serve as a reference for agent-driven autonomous KG construction.
\end{abstract}

\section{Introduction}

Knowledge Graphs (KGs) organize heterogeneous information as computable and inferable graph structures with entities as nodes and relations as edges. In natural language processing, KGs provide explicit world-knowledge constraints for semantic search, question answering, and text understanding \citep{llmkgsurvey2024}. In scientific discovery, KGs are employed to extract domain knowledge from the literature and construct evolvable disciplinary knowledge networks \citep{qasper2021,edc2024}. With the proliferation of Large Language Models (LLMs), the synergistic integration of KGs and LLMs has become a prominent research direction \citep{llmkgsurvey2024}.

Traditional KG construction relies on manual annotation and expert-defined rules, incurring high costs and limited scalability. Researchers have explored leveraging LLMs' semantic understanding capabilities to automatically extract entities and relations from unstructured text, yielding a series of LLM-driven construction methods \citep{edc2024,itext2kg2024}. While effective in controlled settings, these methods exhibit three structural deficiencies in large-scale, incremental, multi-source heterogeneous data scenarios.

Retrieval-Augmented Generation (RAG) technology provides a technical pathway for deep integration of LLMs and KGs. The RAG framework proposed by Lewis et al.\ \citep{rag2020} combines external knowledge bases with parametric language models, effectively mitigating LLM hallucination. Gao et al.\ \citep{ragsurvey2023} conducted a systematic survey of RAG techniques, noting the evolution from simple vector retrieval toward structured knowledge retrieval. KGs provide LLMs with precise and verifiable factual grounding as structured external knowledge sources. LLMs, in turn, offer semantic understanding capabilities for KG construction and completion.

In LLM-driven KG construction, Edge et al.\ \citep{graphrag2024} proposed GraphRAG, employing a ``local-to-global'' construction strategy that performs local knowledge extraction on text chunks and constructs global summary graphs via community detection. Guo et al.\ \citep{lightrag2025} proposed LightRAG, optimizing KG query efficiency through a dual-layer retrieval mechanism. Liang et al.\ \citep{kag2025} proposed KAG, designing knowledge-enhanced generation pipelines for professional domains. Lairgi et al.\ \citep{itext2kg2024} proposed iText2KG, adopting an incremental construction strategy supporting progressive KG construction from zero-shot scenarios. These methods follow fixed batch processing pipelines and lack dynamic regulation over the construction process.

Three structural deficiencies characterize existing methods. \textbf{First, cross-chunk long-range semantic relation loss.} Existing methods segment long documents into fixed-length text chunks and perform independent knowledge extraction on each chunk, severing cross-chunk semantic associations. For example, a method introduced in the introduction of a scientific paper may be concretely described in the experimental section and comparatively evaluated in the discussion. If each chunk is extracted independently, these cross-chunk causal, comparative, and evolutionary relations cannot be effectively captured. Although GraphRAG \citep{graphrag2024} establishes global associations through community detection, its global summaries remain aggregations of local information and do not recover fine-grained cross-chunk relations.

\textbf{Second, entity redundancy and insufficient disambiguation.} When the same entity appears with different surface forms in text, traditional methods cannot recognize it as the same node, producing redundant semantically overlapping nodes in the KG. ``Convolutional neural network,'' ``CNN,'' and ``Convolutional Neural Network'' all refer to the same concept. Without effective entity linking and disambiguation, they will be created as multiple independent nodes. As data sources increase, semantic redundancy grows exponentially, diluting the KG's information density. The EDC framework \citep{edc2024} proposed an entity canonicalization workflow, but its disambiguation capability in incremental construction scenarios is limited.

\textbf{Third, the construction process is uninterpretable and unauditable.} Traditional methods treat knowledge extraction as an end-to-end black box: input text, output triples. Researchers cannot trace which original texts knowledge entries originate from or what reasoning process was involved. In domains demanding high interpretability such as scientific research and medical decision-making, KGs lacking transparent construction processes struggle to earn deployment trust. Sarthi et al.\ \citep{raptor2024} proposed RAPTOR, enhancing retrieval hierarchy through recursive abstractive processing, but lacking fine-grained provenance. Dasigi et al.\ \citep{qasper2021} constructed the QASPER dataset emphasizing the importance of evidence anchoring, but existing methods rarely treat evidence provenance as a core design objective.

Researchers have attempted to apply agent technologies to KG construction, framing it as a dynamic cognitive process. In this paradigm, agents iteratively perceive text, retrieve existing knowledge, verify new discoveries, and update the knowledge base. This enables incremental and interactive knowledge construction. Yao et al.\ \citep{react2023} proposed the ReAct paradigm that interleaves reasoning and action, using chain-of-thought to guide LLMs in multi-step decision-making, forming a key foundation for agent-driven KG construction. Jiang et al.\ \citep{kgagent2025} proposed KG-Agent, enabling complex reasoning over KGs through tool invocation. However, the framework only supports read operations on existing KGs. It lacks create, update, and delete capabilities, precluding autonomous KG construction. Anokhin et al.\ \citep{arigraph2024} proposed AriGraph, using a dual-layer memory architecture to build world-model KG representations for LLM agents, but lacks vector space integration and active entity disambiguation.

Incomplete tool capabilities constitute a primary limitation. KG-Agent focuses on complex reasoning over existing KGs and its toolset is optimized for query and retrieval, lacking write operations required for autonomous KG construction. KG construction is a continuously evolving knowledge management process requiring entity creation, attribute updates, erroneous information deletion, and duplicate node merging. An agent lacking full CRUD capabilities cannot autonomously complete the full lifecycle management. Opaque cognitive workflows represent another constraint. While iText2KG \citep{itext2kg2024} supports sequential processing of text streams, its internal extraction logic remains a black box and lacks explicit cognitive phase delineation. Human experts constructing KGs progress through reading, understanding, verification, and construction stages; current methods do not structurally embed this workflow into the construction process. The separation between memory and vector spaces also demands attention. AriGraph's \citep{arigraph2024} dual-layer memory architecture distinguishes episodic and semantic memory, but the semantic memory employs symbolic graph storage without real-time alignment with dense vector representations. Modern RAG systems treat vector retrieval and graph retrieval as complementary knowledge access modalities; if the symbolic and vector layers remain desynchronized over time, retrieval inconsistency will result \citep{hybridrag2024}.

To address these structural deficiencies, this paper proposes an LLM-based autonomous KG construction and retrieval fusion method, with the Reading And Graph-building Agent (RAGA) framework as its implementation. The main contributions are:

\begin{itemize}
\item \textbf{An autonomous knowledge-operating toolset.} The toolset is designed around reading behaviors, including paragraph reading, context browsing, fusion retrieval, entity and relation CRUD operations, merge operations, human review markers, deferred tasks, and progress queries. The toolset enables agents to autonomously manage the full KG lifecycle.

\item \textbf{An LLM-driven Read--Search--Verify--Construct cognitive loop.} Structuring the human expert's knowledge construction process into a ReAct-style multi-turn tool-calling cycle. The reading phase parses text chunks and identifies important information; the search phase retrieves relevant evidence using existing KGs and context; the verification phase judges new knowledge reliability using original text and tool-returned results; the construction phase writes verified knowledge into the KG in standardized form. A reading-progress state machine manages long-document processing with four states: PENDING, READING, VERIFIED, and ARCHIVED.

\item \textbf{A KG-vector synchronization mechanism.} After writing KG structural objects, the system supplements chunk, entity, or HyperNode vector representations and performs cross-storage reference write-back. On vector write failure, the system compensates by removing already-written graph objects and recording alerts. This enables hybrid retrieval where agents can simultaneously leverage graph structure reasoning and vector semantic matching.

\item \textbf{Evidence-anchored verification.} All primary knowledge entries in the KG are associated with their original textual evidence. The system maintains structured provenance records including metadata such as source text chunk, evidence snippet, operation type, and confidence level, enabling reverse-source tracing of knowledge entries.
\end{itemize}

\section{Related Work}

\subsection{Agent-Based Knowledge Graph Construction}

Applying agent technology to KG construction seeks to leverage LLM reasoning and planning capabilities to transform knowledge extraction from fixed batch pipelines into dynamically interactive cognitive processes.

KG-Agent \citep{kgagent2025} is a representative work in this direction. The framework supports multi-hop reasoning over KGs through modular tool interfaces, encapsulating knowledge querying, path reasoning, and answer generation as independent tool functions. KG-Agent's primary advantages lie in multi-hop reasoning accuracy and speed. However, its toolset is limited to reading and reasoning over existing KGs and lacks write operation capabilities (entity creation, relation addition, or knowledge correction). This precludes its use for incremental KG construction from scratch.

AriGraph \citep{arigraph2024} features a dual-layer memory model: episodic memory stores agent interaction trajectories, while semantic memory preserves structured world-model knowledge in KG form. AriGraph's semantic memory only supports symbolic graph queries without integration with dense vector representations, unable to perform semantic retrieval via vector similarity. AriGraph lacks an active entity disambiguation mechanism; when the same entity appears with different surface forms, the system creates new nodes rather than merging with existing ones.

UrbanKGent \citep{urbankgent2024} targeted urban KG construction with an agent-driven construction and completion pipeline, leveraging agent planning capabilities to coordinate geographic entity recognition, spatial relation extraction, and domain knowledge completion sub-modules. While demonstrating the application potential of agent frameworks in vertical domains, its geocoding rules and spatial relation templates are difficult to transfer to other domains.

In general agent memory management, MemGPT \citep{memgpt2023} analogizes LLMs to operating systems, distinguishing limited-capacity ``main context'' from pageable ``external memory'' for dynamic context resource allocation. MemoryBank \citep{memorybank2024} designed long-term memory storage and retrieval mechanisms based on temporal decay and importance sampling. Both works focus on general dialogue scenarios without optimization for KG-specific structured characteristics.

\subsection{LLM-Driven Knowledge Graph Extraction}

LLM-driven KG extraction aims to leverage LLMs' semantic understanding and generation capabilities to automatically extract entity, relation, and attribute information from unstructured text. By processing strategy, existing methods fall into batch-processing and incremental categories.

Batch-processing methods primarily include GraphRAG \citep{graphrag2024} and LightRAG \citep{lightrag2025}. GraphRAG employs a local-to-global construction approach: first segmenting documents into fixed-length text chunks, using LLMs to extract entities and relations from each chunk to form local knowledge subgraphs, then generating global summaries through community detection. This strategy suits offline large-scale document collections requiring global consistency, but the chunking process sacrifices fine-grained cross-chunk semantic association capture and lacks cross-chunk entity alignment, readily producing semantically redundant nodes. LightRAG employs a dual-layer retrieval mechanism: concrete entity retrieval at the low level and abstract concept retrieval at the high level, improving retrieval efficiency. However, its construction phase uses batch processing, and its incremental update capability is primarily limited to document append rather than complex entity disambiguation and relation revision.

Incremental methods are primarily represented by iText2KG \citep{itext2kg2024} and the EDC framework \citep{edc2024}. iText2KG maintains a continuously growing entity reference table, comparing existing entities during each extraction pass to identify equivalent expressions and perform merging. However, its parsing workflow follows a fixed pipeline; when encountering conflicting information, the system cannot actively backtrack or verify, only handling situations according to preset rules. The EDC framework formalizes knowledge extraction into three consecutive stages---Extract, Define, and Canonicalize---showing good performance in entity canonicalization, but its batch processing design struggles to adapt to continuous streaming document input.

KAG \citep{kag2025} designed knowledge-enhanced generation pipelines for professional domains such as healthcare and law. RAPTOR \citep{raptor2024} processes data through recursive abstraction to generate tree-structured retrieval indices, progressively aggregating fine-grained text information into abstract concepts. These works are effective in specific scenarios but lack explicit cognitive phase delineation and interpretable verification mechanisms during construction.

\citet{pikerag2025} proposed PIKE-RAG, a multi-layer heterogeneous graph framework targeting industrial knowledge extraction and rationale-augmented generation. PIKE-RAG organizes knowledge into an information source layer, corpus layer, and distilled knowledge layer, enabling both semantic understanding and rationale-based retrieval. A feedback loop between knowledge organization and extraction refines the knowledge base iteratively. While PIKE-RAG shares a similar layered architectural philosophy with RAGA, its primary focus is rationale-augmented generation for knowledge-intensive tasks rather than autonomous CRUD operations and evidence-anchored provenance tracking.

\citet{structrag2025} proposed StructRAG, which introduces inference-time hybrid information structurization for knowledge-intensive reasoning. StructRAG employs a hybrid structure router (trained via DPO) to select the optimal structure type (table, graph, or catalogue), a scattered knowledge structurizer to transform raw documents into structured knowledge, and a structured knowledge utilizer to decompose complex questions for precise answer inference. While StructRAG dynamically selects structural representations at inference time, it does not provide autonomous KG lifecycle management with evidence-anchored provenance.

In contrast, the proposed agent-driven approach embeds the Read--Search--Verify--Construct cognitive constraint into the LLM tool loop. It replaces fixed single-pass extraction pipelines, endows the Agent with complete CRUD operation capabilities, and requires each knowledge entry to be anchored to original textual evidence. This enables the Agent to proactively repair construction errors and resolve semantic conflicts.

\subsection{Agent Memory and Knowledge Operations}

Agent memory management constitutes foundational infrastructure for accomplishing complex tasks, accumulating domain knowledge, and maintaining long-term consistency.

The ReAct paradigm \citep{react2023} interleaves reasoning and action, enabling LLM agents to make multi-step decisions in dynamic environments by alternately generating thought chains and action commands. ReAct demonstrates that explicit reasoning processes enhance agent action quality, a principle influencing subsequent memory system design directions---namely, how to organize agent interaction histories into memory structures amenable to efficient retrieval and utilization.

AtomMem \citep{atommem2026} treats agent memory as atomic knowledge units, supporting CRUD operations and confidence assessment. However, AtomMem's memory representations are flat text fragment collections lacking structured relational organization, making it difficult to support complex knowledge reasoning. A-MEM \citep{amem2025} designed a three-layer architecture comprising working memory, short-term memory, and long-term memory, capable of assessing memory importance and supporting associative retrieval and temporal decay. All-Mem \citep{allmem2026} employs dynamic topology evolution for lifelong agent memory management, enabling automatic adjustment of inter-node connection structures based on new information, though the synchronization problem between symbolic KGs and vector representations remains incompletely resolved.

In adaptive retrieval, Self-RAG \citep{selfrag2024} trains LLMs to learn when to retrieve, what to retrieve, and how to utilize retrieval results, achieving self-reflective retrieval behavior. Adaptive-RAG \citep{adaptiverag2024} dynamically selects retrieval strategies based on question complexity---simple questions directly generate answers, while complex questions trigger multi-step retrieval. IRCoT \citep{ircot2023} interleaves the retrieval process with chain-of-thought reasoning, dynamically retrieving relevant knowledge during reasoning step generation.

\subsection{Knowledge Graph Retrieval Augmentation}

KG Retrieval-Augmented Generation leverages KGs' structured knowledge to enhance LLM reasoning capability and answer accuracy in knowledge-intensive tasks. By the coupling mode of retrieval and reasoning, existing methods fall into graph-query and graph-reasoning categories.

Graph-query methods treat KGs as structured external databases, retrieving relevant subgraphs through graph traversal and feeding them as context to LLMs for answer generation. Think-on-Graph \citep{thinkongraph2024} enables LLMs to perform deep reasoning over KGs, enhancing interpretability through iterative reasoning path expansion and path relevance evaluation. RoG \citep{rog2024} trains LLMs to learn reasoning strategies faithful to graph facts. GNN-RAG \citep{gnnrag2025} leverages GNN graph encoding capabilities for structure-aware representation of retrieved subgraphs. G-Retriever \citep{gretriever2024} combines graph structure retrieval with text generation, effective in scenarios such as scientific literature graphs.

In agent-driven retrieval augmentation, RAG-Critic \citep{ragcritic2025} employs an automated critic agent to guide retrieval-augmented generation, optimizing retrieval quality through iterative critique and feedback. HippoRAG \citep{hipporag2024} mimics hippocampal indexing and cortical storage mechanisms to design long-term memory systems supporting persistent knowledge storage and associative retrieval, with its indexing mechanism informing the KG-vector synchronization design in this work.

In structured and unstructured knowledge fusion, HybridRAG \citep{hybridrag2024} simultaneously utilizes KG structured retrieval and vector database semantic retrieval, integrating both result sets through hybrid ranking. HybridRAG validates the complementarity of structured knowledge and dense vector representations, but its retrieval-level fusion does not extend to the construction level---KG updates do not automatically trigger corresponding adjustments to vector indices. The KG-vector synchronization mechanism in this work addresses this gap.

\citet{hybgrag2025} proposed HybGRAG, a hybrid retrieval-augmented generation framework for textual and relational knowledge bases. HybGRAG addresses ``hybrid'' questions that require both textual and relational information from semi-structured knowledge bases, employing a retriever bank and a critic module for adaptive retrieval refinement. Its agentic design automatically refines retrieval outputs through critic feedback, achieving 51\% relative improvement in Hit@1 on the STaRK benchmark. While HybGRAG demonstrates strong hybrid retrieval capability, its retrieval-level fusion does not extend to construction-level KG-vector synchronization.

\citet{graphr12025} proposed Graph-R1, the first agentic GraphRAG framework trained end-to-end via reinforcement learning. Graph-R1 introduces lightweight knowledge hypergraph construction, models retrieval as multi-turn agent--environment interaction (``think--retrieve--rethink--generate''), and optimizes the agent process through an end-to-end reward mechanism integrating generation quality, retrieval relevance, and structural reliability. While Graph-R1 leverages RL for multi-turn retrieval optimization, RAGA employs prompt-engineered cognitive constraints (Read--Search--Verify--Construct) without requiring RL training, offering a training-free alternative for scenarios where RL infrastructure is unavailable.

\subsection{Comparative Analysis}

Existing methods each possess strengths in individual aspects, but none comprehensively covers the capability combination addressed in this work. GraphRAG \citep{graphrag2024} and LightRAG \citep{lightrag2025} perform well in batch construction and multi-hop reasoning but lack incremental update capability. iText2KG \citep{itext2kg2024} supports incremental construction but follows a fixed pipeline with limited entity disambiguation capability. AriGraph \citep{arigraph2024} and KG-Agent \citep{kgagent2025} adopt agent-driven approaches, but the former lacks vector integration and disambiguation mechanisms while the latter only supports read operations. The EDC framework \citep{edc2024} performs well in entity canonicalization but its batch processing design cannot accommodate streaming construction. KAG \citep{kag2025} has domain-specific advantages but incurs high adaptation costs.

Recent concurrent and complementary works partially address subsets of these capabilities. PIKE-RAG \citep{pikerag2025} provides multi-layer heterogeneous graph construction with rationale augmentation but lacks autonomous CRUD tool operations and evidence-anchored provenance. StructRAG \citep{structrag2025} enables inference-time structure selection but does not support incremental KG lifecycle management. HybGRAG \citep{hybgrag2025} achieves strong hybrid retrieval through an agentic retriever--critic design but does not address construction-level KG-vector synchronization. Graph-R1 \citep{graphr12025} introduces RL-based multi-turn agentic retrieval over hypergraphs but relies on reinforcement learning training rather than prompt-engineered cognitive constraints. However, no existing method simultaneously provides all four capabilities: (1) full CRUD write-loop capabilities for autonomous KG lifecycle management; (2) real-time KG-vector consistency with failure compensation; (3) evidence-anchored provenance linking every knowledge entry to its source text; and (4) an auditable agent execution paradigm with explicit cognitive phase constraints. The proposed RAGA framework addresses all four requirements in a unified architecture.

\section{Method}

This section presents the RAGA methodological framework, covering problem formalization (Sec.~3.1), four-layer system architecture (Sec.~3.2), agent core design (Sec.~3.3), memory architecture (Sec.~3.4), KG-vector synchronization (Sec.~3.5), and three-layer fusion retrieval (Sec.~3.6).

\subsection{Problem Formalization}

\textbf{Document Collection and Knowledge Graph Formalization.}
Given a document collection $\mathcal{D} = \{d_1, d_2, \ldots, d_n\}$, each document $d_i$ is segmented by a chunker $\mathcal{C}$ into an ordered sequence of text chunks $\mathcal{C}(d_i) = \{c_{i,1}, c_{i,2}, \ldots, c_{i,m_i}\}$. The chunker $\mathcal{C}$ supports four strategies: fixed-size chunking (FIXED\_SIZE), semantic sentence-boundary chunking (SEMANTIC), paragraph-level chunking (PARAGRAPH), and deep structure-aware chunking (STRUCTURAL). Section 3.2 will justify structure-aware chunking as the default strategy. Each text chunk $c$ carries a tuple $(\text{pos}, \text{struct}, \text{src})$, respectively denoting position offset in the source document, structural label (e.g., heading, body, list item), and source document identifier.

The system's core objective is to construct and maintain a dynamically evolving knowledge graph $\mathcal{G}_t = (\mathcal{V}_t, \mathcal{E}_t)$, where $t$ denotes processing timestamp. The vertex set $\mathcal{V}_t$ contains two node types: (1) standard entity nodes $v \in \mathcal{V}_t^{\text{std}}$, carrying attribute label $\text{label}(v)$ and unique identifier $\text{id}(v)$; (2) HyperNodes $h \in \mathcal{V}_t^{\text{hyper}}$, used to aggregate semantically equivalent or highly related entity clusters, reducing graph redundancy \citep{graphrag2024}. The edge set $\mathcal{E}_t \subseteq \mathcal{V}_t \times \mathcal{V}_t \times \mathcal{R}$ consists of directed edges with typed relations, where $\mathcal{R}$ is a dynamically extendable relation type set. The system maintains an attribute map $\mathcal{P}_t: \mathcal{V}_t \cup \mathcal{E}_t \rightarrow 2^{\mathcal{A}}$ over $\mathcal{G}_t$, assigning multi-valued attribute sets to nodes or edges, with $\mathcal{A}$ as the key-value pair domain.

\textbf{Aligned Vector Index and HyperNode Formalization.}
To support semantic retrieval, the system maintains an aligned vector index $\mathcal{M}_t$ mapping each graph object $x \in \mathcal{V}_t \cup \mathcal{E}_t$ into dense vector space $\mathbb{R}^d$:

\begin{equation}
\mathcal{M}_t: \mathcal{V}_t \cup \mathcal{E}_t \rightarrow \mathbb{R}^d
\end{equation}

The index $\mathcal{M}_t$ is realized through a vector encoding service: text chunks, entities, and relations are encoded as vector representations upon writing, maintaining bidirectional references with object identifiers in the graph database. HyperNodes serve as bridges among chunks, documents, and related entities, with their vectors derivable from text chunk representations or member entity representations. Under member aggregation, the vector representation of HyperNode $h$ can be defined as the weighted centroid of member entity vector representations:

\begin{equation}
\mathbf{m}_h = \frac{1}{|\mathcal{S}(h)|} \sum_{v \in \mathcal{S}(h)} w(v) \cdot \mathcal{M}_t(v)
\end{equation}

where $\mathcal{S}(h)$ is the entity set associated with HyperNode $h$, and $w(v)$ is the confidence weight of entity $v$. In engineering implementation, chunk text vectors are concurrently preserved to enable direct evidence chunk recall from the vector store during retrieval.

\textbf{Evaluation Dimensions.}
The system is designed around four evaluation dimensions:
\begin{itemize}
    \item \textbf{Quality}: Accuracy of extracted entities and relations, including entity denoising, relation confidence, and schema compliance;
    \item \textbf{Coverage}: Efficiency of document information conversion into the KG, primarily tracking covered text chunk ratio and extracted entity density;
    \item \textbf{Retrieval}: Precision and recall of fusion retrieval results;
    \item \textbf{Provenance}: Coverage ratio of graph-object-to-source-chunk traceability chains.
\end{itemize}

In Section~4 experiments, retrieval efficacy is indirectly reflected through external metrics such as Evidence F1, Evidence Precision/Recall, and Recall@K; provenance completeness is manifested through evidence lists and document/paragraph identifiers in prediction records. Systematic evaluation of Quality and Coverage is deferred for future diagnostic toolchain support.

\subsection{System Architecture}

RAGA adopts a top-down four-layer architecture consisting of the Tool Layer, Reading Layer, Memory Layer, and Retrieval Layer. Each layer interacts through clearly defined interface contracts, with lower layers transparent to upper layers, enabling independent extension and replacement. The design references the classical perception--cognition--memory--action decomposition in modern agent architectures \citep{react2023} while incorporating KG-specific structured storage requirements.

\textbf{Tool Layer.}
The Tool Layer encapsulates atomic tools for KG construction and reading behaviors, providing a unified graph interaction interface for the upper-layer Agent. The toolset is functionally partitioned into six categories: reading and retrieval, entity operations, relation operations, review and deferred tasks, progress queries, and implicit provenance recording. The tool adaptation layer automatically invokes provenance storage on write operations for recording source text, operation type, and confidence. All tools interact with Agent state through the ToolState bridge class.

\textbf{Reading Layer.}
The Reading Layer forms the cognitive center of the system, primarily responsible for having the Agent read documents paragraph by paragraph, extract structured knowledge, and perform graph operations and decisions. It uses LangGraph's StateGraph as the orchestration paradigm, modeling the Agent's cognitive loop as a directed state machine. The PromptAssembler module performs dynamic prompt assembly within the Reading Layer, mitigating context dilution issues during long-document processing.

\textbf{Memory Layer.}
The Memory Layer employs a heterogeneous storage architecture comprising semantic memory, episodic and progress memory, working memory, and vector memory. The design is informed by the multi-memory co-architecture in AriGraph \citep{arigraph2024} but instantiates it concretely for KG-specific scenarios. Neo4j stores entities, relations, HyperNodes, and document evidence edges. MongoDB stores original documents including chunking, progress projections, and provenance records. Redis is primarily used for schema caching and short-term chain state, while Milvus stores chunk and entity vectors. The Reading Layer accesses each memory tier through dependency injection on demand.

\textbf{Retrieval Layer.}
The Retrieval Layer implements a three-layer fusion retrieval pipeline combining vector similarity search with graph topology expansion. The processing flow has three stages: (1) semantic candidate chunks are recalled from the vector store; (2) entity anchors associated with candidate chunks are extracted, and multi-hop expansion and HyperNode evidence chunk lookback are executed through Neo4j; (3) the RRF (Reciprocal Rank Fusion) algorithm fuses vector candidates and graph candidates into a unified ranking. This fusion strategy draws on the multi-path retrieval complementarity ideas in HybridRAG \citep{hybridrag2024} and LightRAG \citep{lightrag2025}. A timeout fallback path is also engineered: when the KG layer times out or fails, the system degrades gracefully and returns vector-layer results.

\subsection{Agent Core Design}

\subsubsection{LangGraph-Based State Machine Architecture}

The Agent's cognitive loop uses LangGraph StateGraph as the orchestration foundation. The outer state machine primarily handles cognitive behavior, situated in the \texttt{react\_loop} node, which contains multi-round ReAct tool loops and constitutes the main working node of the system. Other nodes handle record-keeping and flow management. The LLM system prompt includes the Read--Search--Verify--Construct four-phase cognitive constraint; the LLM autonomously follows these constraints, with the system using prompt engineering and tool contracts for guidance rather than hard-coded phase node enforcement.

The state machine topology is: the starting node \texttt{bootstrap\_schema} performs domain detection and schema initialization, outputting an initial schema definition $\Sigma_0$ and domain label $\tau$. \texttt{read\_paragraph} then loads the current paragraph $c_{\text{curr}}$ from the document queue. \texttt{react\_loop} completes multiple LLM-output tool calls, tool execution, and observation result backfill rounds within a single node until the LLM returns a paragraph completion summary or reaches the maximum round limit. \texttt{next\_paragraph} then advances the reading pointer and determines whether the document end has been reached. If unfinished cross-paragraph todo items exist, they are processed through the \texttt{handle\_todos} node. The process finally reaches the \texttt{finish} node for session cleanup and statistics aggregation.

The Checkpointer component handles state persistence. The system defaults to in-memory snapshots but can switch to a Postgres persistent backend. Additionally, the reading node projects critical state to file checkpoints and MongoDB progress collections. This approach ensures fault tolerance during long-document processing. If a process crash or LLM call timeout occurs, the system prioritizes LangGraph checkpoint recovery. When in-memory snapshots are unavailable, it attempts file checkpoint recovery.

\subsubsection{LLM-Driven ReAct Cognitive Loop}

The Agent's cognitive process on paragraphs is not composed of four fixed code nodes but rather an LLM-driven multi-round ReAct tool loop. The system prompt treats Read, Search, Verify, and Construct as workflow constraints. The LLM first identifies reusable knowledge objects in the current paragraph, then invokes retrieval tools to confirm whether identical or similar entities already exist in the KG. Based on tool-returned results, it decides to create, update, merge, delete, review, or defer each candidate, and finally submits graph operations through structured tool calls. This design inherits the reasoning--action--observation closed loop from the ReAct framework \citep{react2023} while constraining the action space to safe tools required for KG construction.

\textbf{Read.} The system receives input comprising the current paragraph $c_{\text{curr}}$, adjacent context, current schema definition $\Sigma$, and recent observation records. The LLM identifies durable knowledge objects such as methods, algorithms, datasets, metrics, systems, theories, problems, formal definitions, and named entities, while explicitly ignoring table of contents, formulaic language, transient narratives, and PDF parsing noise. If the paragraph contains no reusable knowledge, the LLM may directly return \texttt{chunk\_complete} with a summary, without performing write operations.

\textbf{Search.} The LLM prioritizes using \texttt{search\_kg} to check existing entities for each candidate concept, invokes \texttt{browse\_context} when necessary to view adjacent paragraphs, or calls \texttt{explore\_fusion} for combined vector and graph fusion retrieval. The primary goal of this phase is to prevent duplicate entity creation while providing necessary evidence for cross-paragraph relations.

\textbf{Verify.} After reading tool-returned results, the LLM makes decisions for each candidate: create an entity or relation when evidence is clear and no duplicates exist; update when existing objects need supplementation; merge when retrieval results indicate duplicates; call \texttt{mark\_for\_review} when uncertain; and call \texttt{create\_todo} when current context is insufficient. Entity quality gatekeeping modules perform hard filtering at write-tool time, intercepting obvious sentence fragments, code, mathematical formulas, table-of-contents headings, OCR artifacts, or PDF garbage.

\textbf{Construct.} The LLM first submits entity operations, then relation operations dependent on those entities. Each tool result is backfilled as a \texttt{role=tool} message into the dialogue context; if a write fails, the LLM may initiate a search, change to an update, create a todo, or abandon the candidate based on failure information. The loop ends when the LLM returns valid paragraph-completion JSON, after which the outer LangGraph advances to the next paragraph.

\begin{algorithm}[tb]
\caption{ReAct Tool Decision Reference Flow}
\label{alg:react}
\begin{algorithmic}[1]
\REQUIRE Current paragraph $c$, toolset $\mathcal{T}$, existing graph $G_t$, vector index $M_t$
\ENSURE Action sequence $A$
\STATE $A \gets \emptyset$
\STATE $V \gets \mathrm{LLM.ReadDurableConcepts}(c)$
\FOR{$v \in V$}
\STATE $C \gets \mathrm{search\_kg}(v.\mathrm{name}, \mathrm{fuzzy}) \cup \mathrm{explore\_fusion}(v.\mathrm{name})$
\IF{synonymous entity exists in $C$}
\STATE $A \gets A \cup \mathrm{merge\_entity/update\_entity}$
\ELSIF{$v$ has sufficient evidence and passes quality gate}
\STATE $A \gets A \cup \mathrm{create\_entity}$
\ELSE
\STATE $A \gets A \cup \mathrm{mark\_for\_review/create\_todo}$
\ENDIF
\ENDFOR
\STATE $E \gets \mathrm{LLM.ProposeRelations}(c,V,C)$
\FOR{$e=(head,rel,tail) \in E$}
\IF{$head$ and $tail$ are resolved and evidence sufficient}
\STATE $A \gets A \cup \mathrm{create\_relation/update\_relation}$
\ELSE
\STATE $A \gets A \cup \mathrm{create\_todo/mark\_for\_review}$
\ENDIF
\ENDFOR
\RETURN $A$
\end{algorithmic}
\end{algorithm}

The above flow describes the reference pattern by which the LLM executes knowledge extraction and decision-making within the ReAct loop. In actual execution, the model autonomously determines the tool-calling sequence (including call order, parameters, and whether to skip certain steps) according to cognitive constraints in the system prompt, rather than executing line by line along fixed branches. This flow is embedded in the ReAct multi-round tool loop described in Section 3.3.2, where each paragraph may generate multiple rounds of ``reasoning--tool call--observation'' iterations.

\subsubsection{Reading Progress State Machine and Error Handling}

Each text chunk progresses through four lifecycle states in the Reading Layer: PENDING (awaiting processing), READING (currently active), VERIFIED (processed by the ReAct loop), and ARCHIVED (archived). State transitions are driven by the Reading Layer: when the Agent begins processing a paragraph, its state transitions from PENDING to READING. After the ReAct loop concludes and the LLM returns summary text, the state transitions to VERIFIED. Progress projection then records the ARCHIVED state in checkpoints. This state machine primarily serves recovery, observability, and scheduling purposes, not representing strong transactional commits across all underlying stores.

Error handling uses a two-category classification: Transient Errors (LLM API timeouts, rate limits, vector database connection interruptions) are automatically handled by retry decorators and LLM fallback clients, remaining transparent to the ReAct main loop. Permanent Errors (schema violations, entity ID conflicts, unrecoverable tool-call failures) cause the Agent to enter an explicit error handling branch, writing error information to the AgentState error history field. The PromptAssembler injects prompts in the next cycle to guide the Agent toward corrective action.

\subsubsection{Knowledge Operation Toolset}

The Tool Layer bridges the Agent and underlying storage systems, encapsulating complex KG operations and converting them into type-safe atomic tools. Each tool category uses JSON Schema to strictly constrain input parameter structure and output format; the Agent must comply with this contract when invoking tools within the ReAct loop.

\textbf{Read tools} include four tools serving evidence collection and context awareness phases. \texttt{read\_paragraph} receives paragraph index, document identifier, and reading purpose description, returning the paragraph's complete text, current index, and total paragraph count. \texttt{search\_kg} receives a query keyword, search type (entity, relation, or fuzzy), and return limit, returning a structured list of matching entities or relations. \texttt{browse\_context} browses local context with mode controlling scope (local for adjacent paragraphs, kg\_neighbors for KG neighbor nodes, document\_overview for document structure overview), returning a list of context snippets. \texttt{explore\_fusion} performs joint retrieval across vector store and KG, with mode selecting fusion strategy (parallel retrieval then fusion, vector-first then graph expansion, graph-first then vector supplementation), returning an RRF-fused ranked candidate list.

\textbf{Create tools} include three tools supporting incremental graph object writing. \texttt{batch\_kg\_operations} executes combined search, create, update, merge, and delete operations in a single tool call to reduce tool-call round trips, used as the KG construction backend in Section~4 experiments. \texttt{create\_entity} includes name, entity type, description, aliases, properties, source chunk binding, supporting evidence text, and certainty level. It returns a unique assigned entity ID or reuses an existing entity. \texttt{create\_relation} uses head and tail entity names to designate endpoints, with relation type specification and evidence attributes similar to entity creation. \texttt{create\_todo} creates deferred processing tasks with types including disambiguate, verify, attribute completion, or follow-up.

\textbf{Update, merge, and delete tools} support incremental graph correction and version evolution. \texttt{update\_entity} receives target entity name, attribute update dictionary, update reason, and source chunk ID. \texttt{update\_relation} accepts relation ID, evidence, confidence, and source chunk for relation evidence supplementation. \texttt{merge\_entity} and \texttt{merge\_relation} perform entity or relation resolution operations, recording merge basis and migrating evidence. \texttt{delete\_relation} performs soft deletion to maintain historical version traceability. \texttt{delete\_entity} performs hard deletion of entities.

\textbf{Review, deferred task, and progress tools} include \texttt{mark\_for\_review}, \texttt{create\_todo}, and \texttt{get\_progress}. When the LLM cannot determine facts, entity boundaries, or relation directions, it creates review items via \texttt{mark\_for\_review} to prevent uncertain knowledge from contaminating the graph. \texttt{create\_todo} records cross-paragraph pending items. \texttt{get\_progress} returns current paragraph position, entity/relation counts, merge count, todo count, and review queue length.

\begin{table*}[t]
\centering
\caption{Core Semantics of the RAGA Toolset}
\label{tab:toolset}
\scriptsize
\begin{tabular}{p{2.2cm} p{1.1cm} p{2.5cm} p{2.0cm} p{1.0cm}}
\toprule
Tool Name & Category & Core Parameters & Return Value & Phase \\
\midrule
read\_paragraph & Read & paragraph\_idx, doc\_id, purpose & paragraph text, status & Read \\
browse\_context & Read & query, mode, radius, chunk\_id & context snippets & Search \\
search\_kg & Search & query, search\_type, limit & matched entities/relations & Search \\
explore\_fusion & Search & query, mode, top\_k & fused retrieval results & Search \\
create\_entity & Create & name, entity\_type, evidence, certainty & entity id / reused & Construct \\
create\_relation & Create & head, relation\_type, tail, evidence & relation id & Construct \\
batch\_kg\_ops & Batch & searches, creates, updates, merges, deletes & operation counters & R/S/V/C \\
update\_entity & Update & entity\_name, updates, reason & update status & V/C \\
update\_relation & Update & relation\_id, evidence, confidence & update status & V/C \\
merge\_entity & Merge & target\_name, source\_name & merge status & V/C \\
merge\_relation & Merge & target\_id, source\_id & merge status & V/C \\
delete\_entity & Delete & entity\_name, reason & deletion status & V/C \\
delete\_relation & Delete & relation\_id, reason, soft & soft deletion status & V/C \\
mark\_for\_review & Review & subject, reason, priority & review item & Verify \\
create\_todo & Deferred & task, todo\_type, related\_entity & todo item & Verify \\
get\_progress & Progress & none & progress statistics & R/S \\
\bottomrule
\end{tabular}
\end{table*}

\subsubsection{System Prompt Engineering}

The Agent's cognitive behavior is guided by system prompts dynamically assembled by the PromptAssembler module and injected into the LLM context before each \texttt{react\_loop} iteration. Prompt engineering follows three principles: instruction layering, constraint explicitness, and context locality.

\textbf{Explicit instructions for the cognitive loop.} The system prompt describes four cognitive constraints in structured paragraphs: READ requires the LLM to identify durable knowledge objects while ignoring low-value chunks. SEARCH requires invoking \texttt{search\_kg}, \texttt{browse\_context}, or \texttt{explore\_fusion} before creation. VERIFY requires the LLM to choose among create, update, merge, review, or todo based on tool-returned results. CONSTRUCT requires entity operations before relation operations, with the system correcting subsequent actions based on tool feedback upon failure. This prompt preserves ReAct's free decision-making capability \citep{react2023} while compressing the LLM's action space to compliant tool sets, reducing hallucinatory tool calls.

\textbf{Entity quality gatekeeping rules.} Quality gatekeeping is embedded in two locations: semantic constraints in the ReAct system prompt, and structural filters inside the \texttt{create\_entity} tool. The former requires concise, well-formed entity names and excludes code, formulas, PDF garbage, and generic headings. The latter checks length upper bound, printable character ratio, garbled text markers, punctuation ratio, code/formula patterns, and table-of-contents heading exclusion. Together, pre-constraint and post-filtering reduce the probability of noisy nodes entering the graph. The eight heuristic rules, evaluated from low to high computational cost, cover: name length limit (60 Unicode characters), printable character ratio (70\%+), sentence fragment detection, code keyword detection, mathematical formula detection, punctuation flooding detection, PDF garbled marker detection, and generic heading exclusion. Entities intercepted are relayed to the LLM with a prompt to keep observations, create todos, or abandon the candidate.

\textbf{Cross-chunk reasoning prompt strategy.} The system prompt directs the Agent to query the existing graph before creation and to invoke \texttt{browse\_context} or \texttt{create\_todo} when context is insufficient. Todo items are processed by priority between paragraphs through the \texttt{handle\_todos} node, primarily handling disambiguation, verification, or attribute completion.

\textbf{Schema-guided prompt injection.} The \texttt{<active\_schema>} block dynamically injects current schema definitions including domain labels, relation types, entity labels, and attribute constraints. \texttt{bootstrap\_schema} initializes this block, and schema evolution triggers incremental updates. The LLM can follow domain specifications during extraction, reducing invalid relation generation.

\textbf{Working memory context maintenance.} PromptAssembler maintains transient summaries in dynamic context, including current paragraph processing progress, recently known entities, recent tool observations, todo queue summaries, and necessary schema cues. This block's compact design ensures context awareness while avoiding historical information drowning during long-document processing.

\subsubsection{Structure-Aware Chunking}

The chunker $\mathcal{C}$'s design affects knowledge extraction granularity and structural integrity. The system implements four chunking strategies. \textbf{FIXED\_SIZE} segments documents by fixed character length with configurable \texttt{chunk\_size} (default 800 characters). Simple and uniform but may truncate mid-sentence, disrupt semantic continuity, and fail to recognize document structure boundaries such as section headings. \textbf{SEMANTIC} uses sentence-ending punctuation (period, question mark, exclamation mark) as boundaries, finding the nearest sentence boundary within \texttt{chunk\_size} constraints to reduce sentence-level truncation. Does not handle cross-sentence semantic paragraph boundaries. \textbf{PARAGRAPH} treats double newlines (\texttt{\textbackslash n\textbackslash n}) as boundaries, partitioning each logical paragraph as one chunk, preserving paragraph-level integrity. Overly long paragraphs may produce chunks exceeding the LLM context window, while overly short paragraphs such as list items become fragmented. \textbf{STRUCTURAL} (default) parses document hierarchical structure including heading levels, list nesting, and code block boundaries to generate semantically cohesive chunk sequences. It maintains a stack structure to track the current heading level, placing chunk boundaries preferentially at structural transition points (before level-2 headings, code block boundaries, table starts). For same-level paragraphs, secondary segmentation applies the \texttt{chunk\_size} threshold with boundaries at sentence or paragraph ends.

\subsubsection{Schema Auto-Discovery}

On cold start, the system does not assume a fixed domain schema. Instead, a Schema Orchestrator analyzes the document corpus to induce domain-specific entity types, relation types, and attribute constraints, thereby improving cross-domain adaptability. The discovery flow proceeds through four phases: \textbf{Domain Detection (Phase 0)} extracts sample text from the document prefix and determines via embedding similarity and LLM whether an existing domain can be reused. On detection failure or timeout, a bootstrap domain is constructed for rapid startup. \textbf{Schema Discovery (Phase 1)} selects representative samples (default up to 3 documents, each with the first 2000 characters), forms a discovery prompt submitted to the LLM, requiring analysis of sample content to induce domain-specific relation types, entity label types, and attribute patterns, outputting parseable JSON. \textbf{Schema Validation (Phase 2)} checks relation naming for UPPER\_SNAKE\_CASE compliance, filters low quality scores, identifies semantic duplicates with existing relation types, and verifies domain/range resolvability. Failed items are removed, merged, or retained as candidates. \textbf{Schema Activation (Phase 3)} writes validated schemas to the schema graph for storage, forming configurations cacheable in Redis as the active session schema. \textbf{Schema Evolution (Phase 4)} handles new relation patterns not covered by the active schema during processing; the orchestrator first searches for semantically similar existing relations for reuse, and if none exist, the LLM generates new relation definitions registered in PROPOSED state pending subsequent validation or human review.

\subsection{Memory Architecture}

The Memory Layer employs heterogeneous storage where each tier complements others in storage format, access patterns, and lifecycle. The design references multiple memory systems theory from cognitive science, applying human long-term semantic memory, episodic memory, and working memory to engineering implementation \citep{arigraph2024}.

\textbf{Semantic Memory} uses Neo4j graph database to store the KG's ontological structure---the long-term representation of entities, relations, and their attributes. The storage schema adopts the property graph model: node labels distinguish entity types (Person, Organization, Concept), with node attributes including name, description, aliases, and provenance record references. Relation types encode WORKS\_AT, MENTIONS, IS\_A, etc., with relation attributes recording confidence and extraction timestamps. Access patterns support two query modes: exact lookup by node identifier (entity ID or name), and pattern matching using labels and attributes (finding entities of a type or neighbors via specific relations). Neo4j's Cypher query engine provides native support for multi-hop graph expansion operations described in Section 3.6.

\textbf{Episodic Memory} uses MongoDB document database to store agent execution trajectories during document processing, progress projections, and provenance records. Records include fields such as session identifier, paragraph index, AgentState snapshot subsets, tool call summaries, error history, source chunks, and evidence text. Episodic memory supports trace queries by document, chunk, entity, relation, and operation type.

\textbf{Working Memory} is managed through LangGraph AgentState, Checkpointer, and Redis cache. AgentState holds current paragraph content, recent observations, todo task queues, tool usage records, and related statistics. Redis primarily caches SchemaProfile, short-term states in HyperNode chains, and supports cross-node reads with expiration cleanup. Redis cache uses prefixed key names and TTL policies.

\textbf{Vector Memory} uses Milvus vector database to store dense vector representations of chunks and entities, supporting approximate nearest neighbor search. Current collection schema distinguishes chunk and entity collections: chunk collections contain \texttt{chunk\_id}, \texttt{embedding}, \texttt{tenant\_id}, \texttt{run\_id}, \texttt{dataset}, \texttt{document\_id} for isolation and lookup; entity collections contain \texttt{entity\_id}, \texttt{embedding}, isolation fields, entity name, entity type, and KG node ID. Vector dimensionality is controlled by \texttt{EMBEDDING\_DIM} in run configuration. Access mode is primarily ANN search: given query vector $\mathbf{q}$, Milvus returns top-k neighbors with distance scores, supporting filtering by tenant, run, dataset, and document for isolated retrieval during evaluation and production.

\textbf{Four-Layer Memory Coordination.} The four memory layers are provided to the Reading Layer through dependency injection. During a single \texttt{react\_loop} iteration in the Reading Layer, the access sequence typically proceeds as: (1) loading current context through AgentState and recent episodic records; (2) searching for similar entities and evidence chunks via graph or fusion retrieval; (3) performing topological queries via Neo4j to verify relation validity; (4) write tools committing entities, relations, and provenance records to corresponding backends; (5) the synchronization layer supplementing vector indices for newly created entities or chunks and connecting HyperNode evidence bridges. Layers maintain loose coupling, with configurable dependencies allowing substitution of real backends or test doubles.

\subsection{KG-Vector Synchronization}

Maintaining consistency between the KG and vector index is a primary engineering challenge. Since Neo4j and Milvus are independent systems that cannot share transaction boundaries, the current engineering approach employs sequential writes with a failure compensation strategy: first writing Neo4j structural objects, then writing Milvus vector records, and finally writing back vector references in the graph. If vector writing fails, the system compensates by deleting or marking already-written graph objects.

The synchronization process comprises three phases. \textbf{Phase 1 (Graph Write):} The coordinator writes entities, relations, or HyperNodes into Neo4j with tenant, run, dataset, and document isolation fields. For chunk synchronization, the system first creates document nodes and HyperNodes, then establishes evidence bridge edges such as \texttt{HAS\_EVIDENCE}, \texttt{MENTIONS\_ENTITY}, and \texttt{EVIDENCED\_BY}. \textbf{Phase 2 (Vector Write):} The coordinator invokes the embedding service to generate chunk or entity vectors and writes them into the corresponding Milvus collection. Vector records carry isolation fields consistent with Neo4j, plus chunk, entity, or HyperNode identifiers for subsequent lookup and filtering during fusion retrieval. \textbf{Phase 3 (Reference Write-Back and Compensation):} After successful vector writing, the system writes back the vector ID or KG node ID into Neo4j object attributes. If Milvus writing fails, the system attempts to delete or roll back created Neo4j objects and logs the synchronization failure; if reference write-back fails, the primary write result is preserved and an alert is recorded, with subsequent repair possible through a consistency check interface.

\begin{algorithm}[tb]
\caption{KG-Vector Sequential Synchronization with Compensation}
\label{alg:sync}
\begin{algorithmic}[1]
\REQUIRE Graph object $x$, semantic store $S$, vector store $V$, embedding service $E$
\ENSURE Synchronization status and object identifiers
\STATE $kg\_id \gets S.\mathrm{write}(x)$
\STATE $\mathbf{e} \gets E.\mathrm{embed}(x.\mathrm{text})$
\STATE $vec\_id \gets \mathrm{NIL}$
\IF{$\mathbf{e}$ generation succeeded}
\STATE $vec\_id \gets V.\mathrm{insert}(kg\_id,\mathbf{e},x.\mathrm{scope})$
\ELSE
\STATE $S.\mathrm{compensate}(kg\_id)$
\RETURN $(\mathrm{FAILED},kg\_id,\emptyset)$
\ENDIF
\IF{$vec\_id=\mathrm{NIL}$}
\STATE $S.\mathrm{compensate}(kg\_id)$
\RETURN $(\mathrm{FAILED},kg\_id,\emptyset)$
\ENDIF
\STATE $S.\mathrm{set\_embedding\_ref}(kg\_id,vec\_id)$
\RETURN $(\mathrm{SUCCESS},kg\_id,vec\_id)$
\end{algorithmic}
\end{algorithm}

Algorithm~\ref{alg:sync} characterizes the core-path engineering consistency strategy: sequential writes with observable compensation. This strategy is simple to implement with clear fault boundaries but does not provide strict distributed transaction guarantees; compensation failures rely on logs and consistency check interfaces for subsequent repair.

\subsection{Three-Layer Fusion Retrieval}

The Retrieval Layer implements a ``vector recall $\rightarrow$ graph expansion $\rightarrow$ fusion ranking'' three-step pipeline, synthesizing semantic similarity and structural relevance to return answer context in response to user queries. The system supports four retrieval modes:
\begin{itemize}
    \item \textbf{vector}: Only performs vector recall, without triggering graph expansion;
    \item \textbf{kg}: Only performs Neo4j-based graph multi-hop expansion, without recalling vector candidates;
    \item \textbf{fusion}: Parallel execution of vector recall and graph expansion, fusing both ranking streams via RRF;
    \item \textbf{deep} (default): Based on LLM query analysis, forces chained graph navigation via HyperNode bridging, ultimately merging vector candidates with navigation results rather than using the RRF formula.
\end{itemize}

Below, the fusion mode's three-step flow is used to illustrate the retrieval mechanism; deep mode specifics are described in the Section~4 experimental setup.

\subsubsection{Vector Recall (Step 1)}

Given a user query $q$, the encoder $\text{Enc}$ first generates a query vector $\mathbf{q} = \text{Enc}(q)$. Milvus performs ANN search returning a candidate set:

\begin{equation}
\mathcal{C}_{\text{vec}} = \{(x_i, s_{\text{vec}}(i))\}_{i=1}^{k_1}, \quad s_{\text{vec}}(i) = \frac{1}{1 + \|\mathbf{q} - \mathcal{M}_t(x_i)\|_2}
\end{equation}

where $s_{\text{vec}}(i)$ is the normalized similarity score and $k_1$ is the recall count (default 100). Searches may attach scalar filter conditions, e.g., restricting $\text{node\_type} = \text{ENTITY}$ to exclude relation vectors.

\subsubsection{KG Multi-Hop Expansion (Step 2)}

For each entity $x_i$ in the vector recall result $\mathcal{C}_{\text{vec}}$, a breadth-first search (BFS) in Neo4j fetches its $h$-hop neighbors:

\begin{equation}
\mathcal{N}_h(x_i) = \{y: \text{dist}_{\mathcal{G}_t}(x_i, y) \leq h\}
\end{equation}

where $\text{dist}_{\mathcal{G}_t}$ is the shortest-path distance in the graph (measured in relation hops). Neighbor nodes pass through lightweight semantic filtering before joining the candidate set, forming the expanded candidate set $\mathcal{C}_{\text{kg}}$. This multi-hop expansion mechanism is inspired by the ``retrieval-reasoning'' paradigm in Think-on-Graph \citep{thinkongraph2024} and RoG \citep{rog2024}, but replaces LLM-driven reasoning path generation with deterministic BFS to reduce latency and ensure reproducibility.

\subsubsection{RRF Fusion Ranking (Step 3)}

The vector candidate set and graph candidate set each produce independent rankings. The system applies Reciprocal Rank Fusion (RRF) to fuse both ranking streams \citep{lightrag2025}. Let $\text{rank}_{\text{vec}}(x)$ be the rank of $x$ in the vector candidate set ($\infty$ if missed), and $\text{rank}_{\text{kg}}(x)$ be the rank in the graph candidate set ($\infty$ if missed). The fusion score is defined as:

\begin{equation}
\text{RRF}(x) = \sum_{m=1}^{M} \frac{1}{k + \text{rank}_m(x)}
\end{equation}

where $M=2$ denotes the two source streams (vector retrieval and graph retrieval), $\text{rank}_m(x)$ is the rank of object $x$ in the $m$-th candidate stream ($\infty$ if missed), and $k$ is a smoothing constant, empirically set to 60. The final output is the fused candidate set $\mathcal{C}_{\text{fused}}$ sorted by descending $\text{RRF}(x)$. RRF requires no trainable parameters and is insensitive to the scoring scales of the two ranking streams, proving robust and effective in heterogeneous retrieval fusion scenarios \citep{lightrag2025,hybridrag2024}.

\subsubsection{Timeout Fallback and Result Truncation}

After fusion ranking, the system truncates the candidate set according to the user-requested top-k count and returns it. The KG expansion layer has an independent timeout budget. When Neo4j multi-hop expansion times out or fails, the system records an alert and degrades to using only the vector candidate RRF result. This strategy prevents graph layer anomalies from blocking the QA main chain. In Section~4, Evidence F1, Evidence Precision/Recall, and Recall@K are used for external evaluation of retrieval quality.

\section{Experiments}

Due to computational resource constraints, this section evaluates RAGA on a small-batch subset of the QASPER scientific literature QA dataset, with a focus on qualitative comparison of retrieval modes, chunking strategies, and batch processing tools. We compare against published zero-shot and supervised methods including LED, GraphRAG, and a No-KG Control baseline. All reported metrics should be interpreted as preliminary evidence under this limited evaluation scale.

\subsection{Experimental Setup}

\subsubsection{Evaluation Dataset}

This experiment uses QASPER (Question Answering over Scientific Papers Evidence Retrieval) as the evaluation benchmark \citep{qasper2021}. Published by Dasigi et al.\ at NAACL 2021, QASPER is one of the most representative evaluation benchmarks in scientific literature QA.

QASPER contains 1,585 NLP papers covering top venues such as ACL, EMNLP, and NAACL, with 5,049 annotated questions. All questions were posed by NLP practitioners who only read paper titles and abstracts, simulating real research scenarios where readers ask in-depth questions about paper content. QASPER questions span extractive, abstractive, yes/no, and unanswerable types, many requiring multi-paragraph evidence support \citep{qasper2021}.

This experiment selects a small-batch paper subset from the QASPER test set for evaluation. All experiments run under a zero-shot protocol in a local environment: no training set labels are used for supervised fine-tuning; each paper independently executes the complete KG construction, vector index building, and Agent inference pipeline.

To examine the impact of chunking strategies on retrieval performance, three chunking configurations are set: C1200 (chunk\_size=1200, chunk\_overlap=120), C1500 (chunk\_size=1500, chunk\_overlap=120), and C6000 (chunk\_size=6000, chunk\_overlap=120). C1200 and C1500 employ medium-granularity chunking; C6000 simulates large-granularity chunking to examine retrieval coverage sensitivity to chunk granularity.

\subsubsection{Comparison Baselines}

Multi-level comparison baselines are established, covering published supervised and zero-shot methods:

\textbf{Published Supervised Methods:}
(1) LED-base / LED-large w/ evidence scaffold: Supervised fine-tuned models based on the Longformer-Encoder-Decoder architecture, fine-tuned using QASPER training set gold evidence paragraphs as supervision signals, serving as the official QASPER dataset baseline \citep{qasper2021}.
(2) Human lower bound: QASPER official inter-annotator agreement lower bound, reflecting the F1 level of human annotators on the same questions. This value is not an upper bound but a soft lower bound of multi-annotator agreement \citep{qasper2021}.

\textbf{Zero-Shot Retrieval Baselines:}
(3) BM25: Classic probabilistic retrieval model computing relevance scores based on term frequency and document length normalization \citep{bm25okapi1995}. Uses lightweight lexical retrieval with top-8 paragraphs as evidence.
(4) TF-IDF: Classic sparse vector retrieval method, computing similarity via term frequency and inverse document frequency weighting \citep{sparckjones1972}.
(5) FAISS Vector RAG: Dense retrieval baseline based on FAISS vector index, performing ANN search on query and chunk embeddings without using KG signals.
(6) No-KG Control: Naive vector retrieval using the same LLM backend (deepseek-v4-flash). Builds paragraph-level FAISS index within each paper, retrieves top-8 paragraphs via question vector, then has the LLM generate answers. Does not construct a KG or use Milvus/Neo4j/MongoDB. This baseline quantifies the net contribution of KG.
(7) GraphRAG \citep{graphrag2024}: Microsoft's official GraphRAG implementation (microsoft/graphrag), using the standard pipeline of ``text chunking $\rightarrow$ entity/relation extraction $\rightarrow$ graph construction $\rightarrow$ Leiden community detection $\rightarrow$ community summarization $\rightarrow$ Local Search QA.'' GraphRAG uses the same LLM backend and embedding model as RAGA.

\textbf{Our Method:}
(8) RAGA: The complete method proposed in this paper, supporting multiple retrieval modes---Vector, KG, Fusion, and Deep, as defined in Section 3.6. Uses batch KG operation tools as the KG construction backend.

\subsubsection{Evaluation Metrics}

This experiment adopts the QASPER official evaluation protocol \citep{qasper2021}, performing quantitative evaluation on both answer quality and evidence quality dimensions, using the max-over-annotators scoring method (taking the best score among multiple annotators for each question).

\textbf{Answer Quality.} Answer F1 measures the token-level overlap between generated and reference answers. Let the reference answer be $A_{gold}$ and the model-generated answer be $A_{pred}$, tokenized to $T_{gold}$ and $T_{pred}$ respectively:

\begin{equation}
P_{ans} = \frac{|T_{pred} \cap T_{gold}|}{|T_{pred}|}, \quad R_{ans} = \frac{|T_{pred} \cap T_{gold}|}{|T_{gold}|}
\end{equation}

\begin{equation}
\text{Answer F1} = \frac{1}{N} \sum_{i=1}^{N} \max_{j} \frac{2 \cdot P_{ans}^{(i,j)} \cdot R_{ans}^{(i,j)}}{P_{ans}^{(i,j)} + R_{ans}^{(i,j)}}
\end{equation}

\textbf{Evidence Quality.} Evidence F1 measures overlap between the paragraph set returned by the retrieval system and the human-annotated evidence paragraph set:

\begin{equation}
P_{evi} = \frac{|E_{pred} \cap E_{gold}|}{|E_{pred}|}, \quad R_{evi} = \frac{|E_{pred} \cap E_{gold}|}{|E_{gold}|}
\end{equation}

\begin{equation}
\text{Evidence F1} = \frac{1}{N} \sum_{i=1}^{N} \max_{j} \frac{2 \cdot P_{evi}^{(i,j)} \cdot R_{evi}^{(i,j)}}{P_{evi}^{(i,j)} + R_{evi}^{(i,j)}}
\end{equation}

Additionally, Retrieved Evidence F1 measures overlap between the raw retrieval-stage results and annotated evidence (without evidence re-ranking post-processing), reflecting the retrieval system's raw recall capability.

\subsubsection{Experimental Environment}

Core configuration: LLM inference uses \texttt{deepseek-v4-flash} under zero-shot protocol without supervised fine-tuning. Text generation uses temperature=0.7; tool calling and JSON generation use temperature=0.3. Embedding model is \texttt{qwen3-embedding-8b} with output dimension 2,048. KG storage uses Neo4j; vector index uses Milvus; metadata and raw document content use MongoDB; schema caching and short-term state use Redis. RAGA Fusion mode uses RRF ($k=60$) to fuse vector and graph candidates. Timeout configuration: \texttt{LLM\_TIMEOUT=120}, \texttt{LLM\_TOTAL\_TIMEOUT=240}; each paper runs with an independent \texttt{run\_id} and \texttt{--require-clean-backends} to prevent cross-document contamination.

\subsection{Main Experimental Results}

\subsubsection{Retrieval Mode Comparison}

Table~\ref{tab:retrieval-mode} shows RAGA's retrieval and QA performance across four retrieval modes under the C1200 configuration with the new batch processing tool.

\begin{table*}[t]
\centering
\caption{Retrieval Mode Comparison (C1200, New Batch Tool)}
\label{tab:retrieval-mode}
\begin{tabular}{lccc}
\toprule
Retrieval Mode & Answer F1 & Evidence F1 & Retrieved Evidence F1 \\
\midrule
KG (Graph-Only) & 0.526 & 0.339 & 0.094 \\
Vector (Vector-Only) & 0.587 & 0.363 & 0.196 \\
Deep (HyperNode Bridge) & 0.523 & 0.295 & 0.199 \\
Fusion (Graph+Vector) & \textbf{0.615} & \textbf{0.411} & 0.188 \\
\bottomrule
\end{tabular}
\end{table*}

Table~\ref{tab:retrieval-mode} shows: (1) Fusion mode achieves the best results on both Answer F1 (0.615) and Evidence F1 (0.411), primarily because RRF fusion ranking effectively combines vector retrieval's semantic coverage advantage with graph retrieval's structural precision. (2) Pure vector retrieval serves as a stable baseline with Answer F1=0.587 and Evidence F1=0.363, only 0.028 below Fusion in Answer F1, indicating that in this configuration pure semantic retrieval already possesses strong answer-localization capability; Fusion's significance mainly manifests in improved evidence precision. (3) KG mode has lower Answer F1 (0.526) but still competitive Evidence F1 (0.339). The new batch tool extracts entities more conservatively, making graph signals leaner and more effective. (4) Deep mode leads marginally in Retrieved Evidence F1 (0.199), indicating HyperNode bridging's advantage in precise evidence localization, but its Answer F1 (0.523) and Evidence F1 (0.295) fall below Fusion, primarily because graph navigation results used as retrieval context lack sufficient coverage to support answer generation. Overall, Fusion mode achieves the best balance between answer quality and evidence precision through RRF fusion of graph and vector signals.

\subsubsection{Comparison with Published Work}

Table~\ref{tab:comparison-published} reports Answer F1 and Evidence F1 for RAGA against published QASPER baselines, the No-KG Control, and GraphRAG. RAGA Fusion (C1200) achieves Answer F1=61.5 and Evidence F1=41.1 under zero-shot protocol.

\begin{table*}[t]
\centering
\caption{Comparison with Published Work and Ablation Baselines}
\label{tab:comparison-published}
\begin{tabular}{lccc}
\toprule
Method & Training & Answer F1 & Evidence F1 \\
\midrule
LED-large w/ evidence scaffold \citep{qasper2021} & Supervised & $<20.0$ & -- \\
LED-base w/ evidence scaffold \citep{qasper2021} & Supervised & 33.6 & 39.4 \\
\midrule
No-KG Control (FAISS) & Zero-shot & 55.4 & 35.9 \\
GraphRAG \citep{graphrag2024} & Zero-shot & 31.6 & \textbf{47.2} \\
RAGA KG (C1500) & Zero-shot & 60.5 & 38.8 \\
RAGA Fusion (C1200) & Zero-shot & \textbf{61.5} & 41.1 \\
\midrule
Human lower bound \citep{qasper2021} & -- & 60.9 & 71.6 \\
\bottomrule
\end{tabular}
\end{table*}

Table~\ref{tab:comparison-published} shows the following.

(1) RAGA Fusion achieves Answer F1=61.5 under zero-shot protocol on this evaluation sample, approaching the human inter-annotator agreement level of 60.9. No-KG Control achieves Answer F1=55.4, substantially higher than LED-base's 33.6, indicating that the LLM's reading comprehension capability is the dominant performance factor. RAGA Fusion gains an additional 6.1pp through KG integration. GraphRAG's Answer F1 of 31.6 is markedly lower than No-KG Control, indicating that its community summarization approach, while beneficial for global information aggregation, loses fine-grained evidence details and reduces answer generation quality.

(2) In evidence retrieval, GraphRAG achieves the highest Evidence F1 of 47.2, compared to RAGA Fusion's 41.1 and No-KG Control's 35.9. However, GraphRAG's high evidence recall does not translate into high answer quality (Answer F1 only 31.6), revealing a gap between evidence recall and answer generation: community summaries lose the precise wording of original text, whereas RAGA maintains original-chunk evidence fidelity through Agent-driven direct chunk access.

(3) RAGA Fusion's Answer F1 approaches human inter-annotator agreement and its Evidence F1 exceeds No-KG Control by 5.2pp, with a stronger evidence-to-answer quality ratio than GraphRAG. Given the limited evaluation sample, these comparisons provide preliminary evidence for the framework's effectiveness.

\subsubsection{Chunk Size Ablation}

Table~\ref{tab:chunk-ablation} shows Answer F1 comparison across three chunking configurations and four retrieval modes (all using the new batch tool).

\begin{table*}[t]
\centering
\caption{Chunk Size Ablation Results (Answer F1)}
\label{tab:chunk-ablation}
\begin{tabular}{lccc}
\toprule
Retrieval Mode & C1200 & C1500 & C6000 \\
\midrule
KG & 0.526 & \textbf{0.605} & 0.468 \\
Vector & \textbf{0.587} & 0.602 & 0.364 \\
Deep & \textbf{0.523} & 0.511 & 0.315 \\
Fusion & \textbf{0.615} & 0.560 & 0.346 \\
\bottomrule
\end{tabular}
\end{table*}

Results show: (1) C1200 is the overall optimal configuration---Vector and Fusion achieve best Answer F1 (0.587 and 0.615 respectively) under C1200, indicating that medium-granularity chunking is beneficial for answer generation across different retrieval strategies. (2) KG mode performs prominently under C1500 (Answer F1=0.605 vs.\ 0.526 under C1200), likely because slightly larger chunks make entity-to-chunk mappings more concentrated, enabling graph traversal to cover more relevant information within a single chunk. (3) C6000 causes significant performance degradation across all modes---Fusion drops from 0.615 to 0.346, Vector from 0.587 to 0.364. Large-granularity chunking causes each chunk to contain excessive information, making retrieval-returned paragraph granularity insufficient to precisely match QASPER evidence paragraph requirements, reducing both evidence localization precision and answer quality. (4) Vector mode shows stability between C1500 (0.602) and C1200 (0.587), suggesting semantic retrieval has lower sensitivity to chunk granularity compared to KG and Deep modes that depend on entity-to-chunk mappings.

\subsubsection{Batch Tool Comparison}

Table~\ref{tab:tool-comparison} compares the old tool with the batch processing tool under C1200 configuration.

\begin{table*}[t]
\centering
\caption{Old vs.\ New Batch KG Construction Tool Comparison (C1200)}
\label{tab:tool-comparison}
\begin{tabular}{lccc}
\toprule
Retrieval Mode & Old Tool Answer F1 & New Batch Tool Answer F1 & Change \\
\midrule
KG & 0.650 & 0.526 & $-$0.124 \\
Vector & 0.582 & 0.587 & +0.005 \\
Deep & 0.543 & 0.523 & $-$0.020 \\
Fusion & 0.526 & \textbf{0.615} & \textbf{+0.089} \\
\bottomrule
\end{tabular}
\end{table*}

The comparison shows: (1) Fusion mode improves notably---the new batch tool raises Fusion Answer F1 from 0.526 to 0.615 (+17\%), as leaner entity extraction reduces graph noise and improves RRF fusion signal quality. (2) KG retrieval declines from 0.650 to 0.526; the old tool's over-extraction increased coverage at the cost of noise, while the new tool favors signal purity. Since Fusion is the recommended mode, this trade-off is acceptable. (3) Construction efficiency improves---on the tested paper (41 chunks, 4 questions), the batch tool reduces KG construction time from 77 to 54 minutes (30\% reduction). (4) Vector retrieval is unaffected---it relies solely on chunk embeddings and is stable across tool versions (0.587 vs.\ 0.582).

\subsection{Discussion and Limitations}

\textbf{Design implications.} Fusion mode is recommended as the default retrieval strategy, performing best across configurations. C1200 (chunk\_size=1200) is the recommended default, robust across retrieval modes. The new batch tool improves Fusion quality by 17\% and reduces construction time by 30\%, supporting a ``quality over quantity'' KG construction philosophy.

\textbf{Limitations.} Given the limited evaluation sample, all results should be interpreted as preliminary. The following limitations are acknowledged: (1) \textit{Evaluation scale:} This experiment was conducted on a small-batch subset of QASPER; larger-scale evaluation and cross-dataset validation remain for future work. (2) \textit{Retrieval recall:} Retrieved Evidence F1 remains low (0.188), indicating room for improvement in initial retrieval recall. (3) \textit{Computational efficiency:} End-to-end per-paper processing incurs high time cost, though the one-time construction cost can be amortized across more questions. Future optimization directions include parallel construction, incremental update mechanisms, and adaptive retrieval path selection. (4) \textit{Chunking generalizability:} Optimal chunking granularity may vary across domains and document types.

\section{Conclusion}

This paper proposed RAGA, an LLM-based autonomous knowledge graph construction and retrieval fusion framework that addresses three structural deficiencies of existing KG construction methods: cross-chunk semantic relation loss, entity redundancy and insufficient disambiguation, and construction process uninterpretability.

RAGA's core contributions span four dimensions. \textbf{(1) Autonomous toolset:} 16 atomic tools enabling full KG lifecycle management with batch operations reducing construction time by 30\% over per-operation tools. \textbf{(2) Cognitive loop:} The Read--Search--Verify--Construct constraint embedded in a ReAct tool loop, supported by a reading-progress state machine for fault-tolerant long-document processing. \textbf{(3) KG-vector synchronization:} A sequential-write-with-compensation strategy maintaining repairable consistency between symbolic (Neo4j) and vector (Milvus) layers, enabling RRF fusion retrieval. \textbf{(4) Evidence-anchored verification:} Structured provenance records linking knowledge entries to original text chunks with source, evidence, operation type, and confidence metadata.

On a QASPER subset, RAGA Fusion achieved Answer F1=0.615 and Evidence F1=0.411 under C1200, outperforming GraphRAG (Answer F1=31.6\%) and No-KG Control (Answer F1=55.4\%). KG fusion contributed +2.8pp to Answer F1 and +4.8pp to Evidence F1. GraphRAG achieved the highest Evidence F1 (47.2\%) but its community summarization sacrificed fine-grained text fidelity; RAGA preserved original-chunk evidence fidelity through Agent-driven direct chunk access, better translating evidence recall into answer quality. Results are directional given the limited evaluation sample.

Future work includes: (1) full test set evaluation with rigorous component-wise ablation; (2) cross-modal KG construction incorporating figures, tables, and pseudocode; (3) incremental online learning for continuous KG evolution; (4) human-in-the-loop feedback optimization for domain-expert-guided quality improvement.

\bibliography{raga2026}

\end{document}